\documentclass[sigconf,11pt]{acmart}

\usepackage{booktabs} 

\usepackage{algorithm}
\usepackage{algorithmic}
\usepackage{amsmath}
\usepackage{amsfonts}
\usepackage{bbm}
\usepackage{graphicx} 
\usepackage{enumitem}
\usepackage{float}
\usepackage{mathtools}
\usepackage{enumitem}
\usepackage{subcaption}

\def\ci{\perp\!\!\!\perp}

\setcopyright{none}

\acmConference[]{2017 Workshop on Fairness, Accountability, and Transparency in Machine Learning}{}{Halifax, Nova Scotia, Canada} 
\acmYear{2017}

\begin{document}
\title{Causal Falling Rule Lists}

\author{Fulton Wang}
\affiliation{\institution{MIT}}
\email{fultonw@mit.edu}

\author{Cynthia Rudin}
\affiliation{\institution{Duke University}}
\email{cynthia@cs.duke.edu}

\renewcommand{\shortauthors}{Wang and Rudin}

\begin{abstract}
A causal falling rule list (CFRL) is a sequence of if-then rules that specifies heterogeneous treatment effects, where (i) the order of rules determines the treatment effect subgroup a subject belongs to, and (ii) the treatment effect decreases monotonically down the list.  A given CFRL parameterizes a hierarchical bayesian regression model in which the treatment effects are incorporated as parameters, and assumed constant within model-specific subgroups.  We formulate the search for the CFRL best supported by the data as a Bayesian model selection problem, where we perform a search over the space of CFRL models, and approximate the evidence for a given CFRL model using standard variational techniques.  We apply CFRL to a census wage dataset to identify subgroups of differing wage inequalities between men and women.
\end{abstract}

%
%

\maketitle

\section{Introduction}


In identifying heterogeneous treatment effects, the end goal is often to rank subgroups by the treatment effects within, so that those for which a treatment is most effective can be treated first.  
This segmentation of data into regions of differential treatment effect has been of recent interest in social science, medical, and marketing domains \cite{Brodersen2015,Sun2014,cai2011analysis,foster2011subgroup, taddy2014heterogeneous}, precisely so that the relevant subgroups can be given priority treatment.  For example, a drug can be given to the patient group for whom it is most effective, or an ad can be shown to the audience most likely to be swayed by it.  Methods have used tree structures \cite{Su2012,Su2009,athey2015machine,beygelzimer2009offset} to form such treatment effect subgroups.  Rule trees have the benefit of being interpretable in defining the partitions, and (potentially) sparse in the number of partitions.  However, tree based methods suffer from two drawbacks: (i) their training is based on greedy splitting criteria, and (ii) given a partitioning tree, it is still cognitively demanding to perform the downstream decision-making task of ranking subgroups by treatment effect and identifying the logical combination of rules defining each.

To address the shortcomings of past tree-based methods, we introduce \emph{causal falling rule lists}.  A causal falling rule list (CFRL) is a Bayesian model parameterized by a sequence of if-then rules such that (i) the sequence of rules determines which treatment effect subgroup a subject belongs to, and (ii) the treatment effect for each subgroup \emph{decreases monotonically} as one moves down the list.  For example, a CFRL might say that if a person is below 60 years, then they are in the highest treatment effect subgroup, such that administering a drug will result in a 20 unit increase in good cholesterol levels.  Otherwise, if they are regular exercisers, then taking the drug will result in a 15 unit increase in cholesterol level.  Finally, if they satisfy neither of these rules, the drug will result in only a 2 unit increase.  Thus, the hallmark of a given CFRL is that the treatment effect is modelled as being \emph{constant} within a single subgroup and ``falling'' along the subgroups.

We choose to formulate the search for the ``best'' CFRL as a \emph{Bayesian} model selection problem.  Thus for a given CFRL, we place a prior over its parameters and choose the CFRL for which the evidence, the likelihood averaged over the parameter prior, is highest.  This approach uses Bayesian Occam's Razor to penalize overly complex models, and avoids having to perform cross validation to tune the complexity penalty parameters.  The model search process will identify heterogenous treatment effects by identifying the model under which the assumption of ``falling'' constant treatment effects across the subgroups of the given model are most likely.  The special structure of a CFRL addresses the shortcomings of tree-based methods.  Firstly, non-greedy training procedures become feasible when the search space is reduced from that of trees to lists.  Secondly, the monotonicity constraint over treatment effects directly ranks the treatment effect subgroups in the order they should be targeted.  This greatly increases its interpretability, which we believe is its main advantage.

\section{Model}

\subsection{Notation}
In this work, we assume a binary treatment and a dataset of $N$ units indexed by $(n)$, who each have $K$-dimensional covariate vectors $x^{(n)} \in \mathbb{R}^K$.  We use the Rubin potential outcomes framework \cite{rubin1974estimating}, with potential outcomes $Y_1^{(n)},Y_0^{(n)} \in \mathbb{R}$ under treatment and control, respectively, and treatment assignment indicator $T^{(n)} \in \{0,1\}$.  $x,Y_1,Y_0,T$ will refer to the set of covariates, potential outcomes, and treatment assignments for all $N$ units in the training data, collectively.  We denote the observable and unobservable outcome for the $n$-th unit as 
\begin{align}&Y^{(n)}=T^{(n)}Y_1^{(n)} + (1-T^{(n)})Y_0^{(n)},\\ 
&\bar{Y}^{(n)}=(1-T^{(n)})Y_1^{(n)} + T^{(n)}Y_0^{(n)},\ \text{respectively.} \label{eqn:observed}
\end{align}  Similarly $Y$ and $\bar{Y}$ to refer to the observable and unobservable outcomes for the entire data.  According to the Bayesian approach we will assume a distribution $P(Y_1,Y_0,T,\theta|x)$, where $\theta$ refers to unknown parameters of our model.  Given observed data, the goal will be to, for a test sample indexed by $(*)$, under our model, present the posterior of $Y_1^{(*)} - Y_0^{(*)}|x^{(*)}$.  This distribution will depend only on the posterior of $\theta$, whose inference will be our main focus.

\subsection{Assumptions}
We assume that all covariates $x$ and observable outcomes $Y$ are available.  Secondly we assume the Bayesian version of conditional ignorability \cite{rubin1978bayesian}, that under the assumed model $P(Y_1,Y_0,T,\theta|x)$, 
\begin{align}
T \ci \theta, Y_1,Y_0| x
\end{align}
Under this assumption, $P(Y_1, Y_0, T, \theta|x) = P(Y_1, Y_0, \theta|x)$ regardless of what $P(T|x)$ is.  Thus, the modeller need only provide a model of $P(Y_1, Y_0, \theta|x)$, and we provide this model under the factorization $P(Y_1, Y_0| \theta,x) P(\theta|x)$.   

\subsection{Parameterization}
Accordingly, a CFRL is a Bayesian model of $Y_1^{(n)},Y_0^{(n)}$ given $x^{(n)}$.
   A given CFRL is parameterized by the length and sequence of rules in it (and subsequent hyperparameters, specified later): 
\begin{align}
&L \in \mathcal{Z}^+ \ \text{(length of list)},\\  
&c^{(l)}(\cdot) \in C\ \  \ \text{for}\ l=1,\dots,L\ \text{(rules in list)},
\end{align}
where $C$ represents the space of rules, namely that of boolean functions on feature space $\mathbb{R}^K$.  These $L$ rules partition the feature space into $L$ regions within which the treatment effects are assumed constant.  We notate a model by its rule list $\{c^{(l)}(\cdot)\}$ or $M;\{c^{(l)}(\cdot)\}$, omitting dependence on hyperparameters when appropriate.


\subsection{Generative Process}
Under a model with rule sequence $\{c^{(l)}(\cdot)\}$, which we will denote $M;\{c^{(l)}(\cdot)\}$, the subjects are assigned treatment effect subgroup $z^{(n)} \in \{1,\dots,L\}$ according to the logic of a decision list:
\begin{align}
z^{(n)} = \min(l; c^{(l)}(x^{(n)}) = 1, l = 1,\dots,L). \label{z_n}
\end{align}
We will always assume that the last rule, $c^{(L)}(\cdot)$, is a \emph{default rule} that always returns true, so that this $\min$ is well defined.

A separate regression model within each subgroup controls for confounding covariates and models the impact of receiving the treatment on outcome, giving the likelihood:
\begin{align}
&Y_1^{(n)}|x^{(n)} \sim \mathcal{N}(D^{(z^{(n)})} + B^{(z^{(n)})\prime} x^{(n)}, \tfrac{1}{\lambda^{(z^{(n)})}}),\label{eqn:likelihood}\\ 
&Y_0^{(n)}|x^{(n)} \sim \mathcal{N}(B^{(z^{(n)})\prime} x_n, \tfrac{1}{\lambda^{(z^{(n)})}})
\end{align}
depending on parameters
\begin{align}
&B^{(l)} \in \mathbb{R}^K\       \text{(subgroup regression coefficient)},\\ &\lambda^{(l)} \in \mathbb{R}^+\   \text{(subgroup noise precision)},\\
&D^{(l)} \in \mathbb{R}\    \text{(subgroup treatment effect)}
\end{align}
 for $l=1\dots L$, and under the constraint that the treatment effects decrease down the list:
\begin{align}
D^{(l)} &> D^{(l-1)}\  \text{\ for}\ l=1\dots L-1. \label{eqn:mono}
\end{align}


\subsection{Prior}
The joint prior over subgroup treatment effects $D^{(1)},\dots,D^{(L)}$ must respect the monotonicity constraints of Equation (\ref{eqn:mono}).  Thus, we perform the reparameterization 
\begin{align}
D^{(l)} = \sum_{l'=L}^l\delta^{(l')}
\end{align}
and place uniform priors with only support over the positive reals on (all but one of) the $\delta^{(l)}$:
\begin{align}
\delta^{(l)} &\sim \operatorname{Uniform}(0,s_0) \ \ \text{for}\ l=1,\dots,L-1,\end{align}
with $s_0\geq0$, and $\delta^{(L)} \sim \operatorname{Uniform}(r_0,s_0).$  Thus we enforce the monotonicity of Equation (\ref{eqn:mono}) as a ``hard'' constraint that will still be true in the posterior.  For example, in the posterior, $E[D^{(l)}]>E[D^{(l+1)}]$.


We assume each $B^{(l)}$ is written as the concatenation $B^{(l)} = [B^{(l)}_h B^{(l)}_i]$ with $B^{(l)}_h \in \mathbb{R}^{K_h}$, $B^{(l)}_i \in \mathbb{R}^{K_i}$, $K_h+K_i=K$; strength is shared between the $B^{(l)}_h$ through an hierarchical prior, and the $B^{(l)}_i$ are a priori independent:
$\mathbb{R}^+\ni \tau \sim \operatorname{Wishart}(v_0, w_0),
\mathbb{R}^{K_h}\ni m \sim \mathcal{N}(\mathbf{0}_{K_h},(c_0 I_{K_h})^{-1}),
\mathbb{R}^{K_h}\ni B^{(l)}_h \sim \mathcal{N}(m, (\tau I_{K_h})^{-1}),
\mathbb{R}^{K_i}\ni B^{(l)}_i  \sim \mathcal{N}(\mathbf{0}_{K_i}, (u_0 I_{K_i})^{-1}),$
where $\mathbf{0}_K$ denotes the $K$-dimensional 0-vector, $I_K$ denotes the $K$-dimensional identity matrix, and the Wishart distribution is 1-dimensional (a reparameterized Gamma).

Finally, we let 
$\lambda^{(l)} \sim \operatorname{Gamma}(\alpha_0,\beta_0)$
so that the complete set of hyperparameters are
\begin{align}
s_0,v_0,w_0,c_0,u_0,\alpha_0,\beta_0 \in \mathbb{R}^+,\ 
r_0 \in \mathbb{R}. \label{eq:hyperparameters}
\end{align}

\section{Model Selection}
\subsection{Evidence Approximation}
To evaluate a model $M;\{c^{(l)}(\cdot)\}$, we use the evidence
\begin{align}
p(Y;M) &= \int_{\theta} p(Y|\theta;M)p(\theta;M) d\theta,
\end{align}
where $\theta=\{\vec{B},\vec{\delta},\vec{\lambda}, m, \tau\}$
are the latent parameters, where $\vec{B}$ denotes the $L$ parameters $\{B^{(l)}\}$ (likewise for $\vec{\delta},\vec{\lambda}$), and
$\mathcal{H}$ are the hyperparameters of the model as described in Equation (\ref{eq:hyperparameters}).  As this integral is not analytically available, and computationally infeasible to calculate using sampling, we approximate it using a standard variational approach.

\subsection{Model Search}
We perform model selection over models $M;\{c^{(l)}(\cdot)\}$ where each rule of the rule list is assumed to come from a pre-mined set of boolean functions $C$ returned by a frequent item-set mining algorithm.  For this particular work, we used FPGrowth \cite{borgelt2005implementation}, whose input is a binary dataset where each $x$ is a boolean vector, and whose output is a set of subsets of the features of the dataset. For example, $x_2$ might indicate the presence of diabetes, and $x_{15}$ might indicate the presence of hypertension, and a boolean function returned by FPGrowth might return 1 for patients who have diabetes and hypertension.  

We use simulated annealing over model space to maximize model evidence.  The proposal distribution involves choosing one of the following random changes at uniform: swapping two randomly chosen existing rules in the list, replacing a randomly chosen rule in the list with another one not in it, increasing its length by inserting a new rule at a random position, and decreasing its length by deleting a randomly chosen rule.
\begin{figure*}
 \hspace*{-0pt} \begin{tabular}{llllll}
  & Conditions                            &                 & Support &    Effect  & Match \\\hline
 IF         & Occup=prof. specialty AND race=white         & THEN treatment effect is: & 579& \$5.49  &\$4.01\\
 ELSE IF    & Occup=factory AND union=no   & THEN treatment effect is: & 531 & \$3.93 & \$2.14 \\
 ELSE IF    & Occup=sales AND householder=false    & THEN treatment effect is: & 492& \$2.32  &\$1.07 \\
 ELSE IF    & Industry=trade AND householder=false                    & THEN treatment effect is: & 649 & \$2.08 &\$0.40  \\
 ELSE IF    & govt employer=false AND no college educ. & THEN treatment effect is: & 3939 & \$1.86  &\$2.73 \\
 ELSE IF    & Industry=education                     & THEN treatment effect is:  & 255 &\$1.11  &\$1.36 \\
 ELSE       &                             & treatment effect is:     & 1103 & \$0.55  & \$0.75 
 \end{tabular} 
 \caption{Causal falling rule list for treatment effect of being male on hourly wage.} \label{table:wage_model}
 \end{figure*}
 


\vspace*{-15pt}
\section{Simulation Studies}

We show that for simulated data generated by a known CFRL model, our simulated annealing procedure with high probability, recovers the model.
Given observations with arbitrary features, and a collection of rules on those features, we can construct a binary matrix where the rows represent observations and the columns represent rules, and the entry is 1 if the rule applies to that observation and 0 otherwise. 
We generated independent binary rule sets with 100 rules by setting each feature to 1 independently with probability 0.25.

\begin{figure}[h]
\includegraphics[width=1\linewidth]{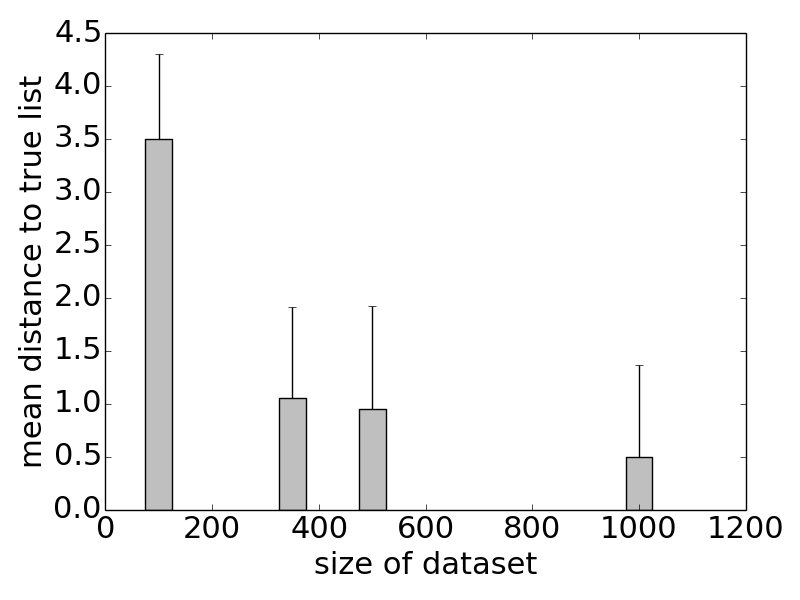}
\caption{Mean distance to true list decreases with increasing sample size.}
\label{fig:sim_study}
\end{figure}

Then, for each $N$, we performed the following procedure 20 times: We generated the random rule matrix, generated a random CFRL of size 6 by selecting 5 rules at random plus the default rule, and, assuming 10 confounding features, for $l=1,\dots,6$, generated pararameters $B^{(l)} \sim \mathcal{N}(0_{10},I_{10})$ and set $\lambda^{(l)}=1$.  For each $n=1,\dots,N$ we generated $x^{(n)} \sim \mathcal{N}(0_{10},I_{10})$, $T^{(n)}$ uniformly from $\{0,1\}$ and simulated $Y^{(n)}$ according to Equation (\ref{eqn:likelihood}) and (\ref{eqn:observed}) to obtain an independent dataset of size $N$.  We ran the simulated annealing procedure to obtain an estimate of the true model, and calculated its edit distance to the true one, and display in Figure \ref{fig:sim_study} the average distance over these 20 replicates.

\begin{figure}[h]
\includegraphics[width=1\linewidth]{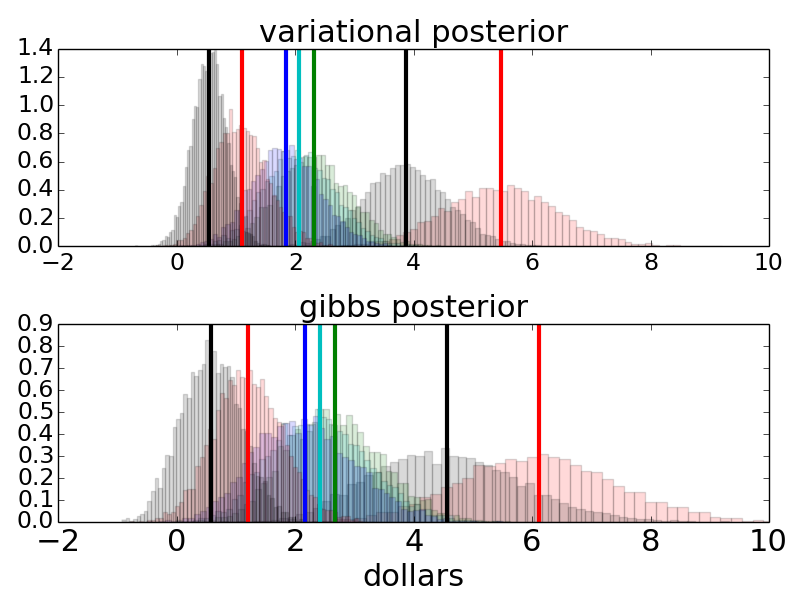}
\caption{Posterior treatment effects for each subgroup as obtained by variational inference and Gibbs sampling. Due to the ``falling'' constraint, the rightmost distribution corresponds to the treatment effect for top subgroup in the list, etc.\label{fig:effect_distribution} }
\end{figure}


\section{Application to Wage Data}
We apply CFRL to a dataset that contains the hourly wage of individuals to assess the treatment effect of gender on hourly wage.  We view the act of changing one's gender from female to male as the treatment.  While this treatment is admitted infeasible, another study \cite{Su2009} also previously studied the effect of gender on wages, using the same dataset.  This dataset was collected in 1995 through the US Census' Current Population Survey \cite{CPS} and after removing individuals who were unemployed or for whom salary data was not available, the dataset retains the salary and gender of 7548 individuals, along with 15 covariates.  The covariates included mostly categorical features such as industry, marital status, union status, education level, as well as 2 scalars: age and weeks worked.  As the FPGrowth rule miner we use accepts binary features only, we discretize the scalar features and group the levels of categorical features manually to obtain $K=54$ binary features so that each $x^{(n)} \in \{0,1\}^{54}$.  We mined all rules with a support of at least 5\% and at most 2 clauses, to obtain a set of 561 rules.  The mean hourly wages was \$8, and rarely above $25$.

We ran simulated annealing for 5000 steps, with a constant temperature of 1, and initializing with a random rule list of length 3.  We display in Figure \ref{table:wage_model} the rules of the model with the highest evidence, as well as the mean posterior treatment effects for each subgroup from variational inference in the column ``Effect''.  In Figure \ref{fig:effect_distribution}, we show in the top panel the posterior variational distributions of the treatment effects for each subgroup.  To compare the variational posterior to the true one, we implemented Gibbs sampling for the model, and show the treatment parameter posterior from 7500 Gibbs steps in the bottom of Figure \ref{fig:effect_distribution}.  

For comparison, we show the treatment effect estimates obtained via propensity score matching using the Matching R package \cite{sekhon2008multivariate} in the column ``Match''.  One notices that the mean posterior treatment effects of our model are inflated relative to those from matching.  However, this is inevitable due to the monotonicity constraint on the joint prior over strata treatment effects, as any sample from the prior will have $D^{(i)}>D^{(i+1)}$.

\bibliographystyle{ACM-Reference-Format}
\bibliography{bib1,frl_bib} 

\end{document}